\RequirePackage{fix-cm}
\documentclass[smallextended]{svjour3}       
\smartqed  
\usepackage{graphicx}
\usepackage{lineno,hyperref}
\usepackage{framed,multirow}

\usepackage{amssymb}
\usepackage{amsmath}
\usepackage{latexsym}
\usepackage{subfig}
\usepackage{float}
\usepackage{enumitem}
\usepackage[numbers]{natbib}
\usepackage[misc]{ifsym}
\usepackage[symbol]{footmisc}
\usepackage{url}
\usepackage{xcolor}
\definecolor{orange}{rgb}{.8,.349,.1}
\definecolor{blue}{rgb}{.1,.1,.8}
\definecolor{green}{rgb}{0.0,0.5,0.0}
\definecolor{green1}{rgb}{0.0, 0.26, 0.15}
\definecolor{pink1}{rgb}{.87,.19,.39}
\definecolor{pink2}{rgb}{0.9, 0.17, 0.31}
\definecolor{pink}{rgb}{.8,.1,.6}
\definecolor{lavender}{rgb}{0.45, 0.31, 0.59}
\definecolor{darkpink}{rgb}{0.53, 0.15, 0.34}
\definecolor{red}{rgb}{0.55, 0.0, 0.0}
\definecolor{red1}{rgb}{0.76, 0.13, 0.28}
\definecolor{violet}{rgb}{0.54, 0.17, 0.89}
\begin{document}

\title{ A Deep and Wide Neural Network-based Model for  Rajasthan Summer Monsoon Rainfall (RSMR) Prediction}



\author{Vikas Bajpai\footnote{\label{corresAuth} The authors contributed equally}\and 
		Anukriti Bansal \footref{corresAuth}
}
\institute{V. Bajpai  \at
              The LNM Institute of Information Technology \\
			  Jaipur, Rajasthan, India \\
              \email{vikas.bajpai87@gmail.com}           
           \and
           A. Bansal \at
              The LNM Institute of Information Technology \\
			  Jaipur, Rajasthan, India \\
			  \email{anukriti1107@gmail.com}
}

\date{Received: date / Accepted: date}
\authorrunning{
Vikas Bajpai \and 
		Anukriti Bansal \and 
}
\maketitle

\begin{abstract}

Importance of monsoon rainfall cannot be ignored as it affects round the year activities ranging from agriculture to industrial. Accurate rainfall estimation and prediction is very helpful in decision making in the sectors of water resource management and agriculture. Due to dynamic nature of monsoon rainfall, it's accurate prediction becomes very challenging task. In this paper, we analyze and evaluate various deep learning approaches such as one dimensional Convolutional Neutral Network, Multi-layer Perceptron and Wide Deep Neural Networks for the prediction of summer monsoon rainfall in Indian state of Rajasthan.For our analysis purpose we have used two different types of datasets for our experiments. From IMD grided dataset, rainfall data of 484 coordinates are selected which lies within the geographical boundaries of Rajasthan. We have also collected rainfall data of 158 rain gauge station from water resources department. The comparison of various algorithms on both these data sets is presented in this paper and it is found that Deep Wide Neural Network based model outperforms the other two approaches.
 
\end{abstract}

\section{Introduction}
\label{sec:introduction}

Understanding of rainfall characteristics is important for a variety of activities including efficient engineering, planning and management of water resources \cite{halbe2013towards, campling2001temporal}. 
In addition to this, rainfall play a major role in balancing of various activities such as, hydrologic cycle,  water availability for terrestrial animals,  agriculture and industrial processes. Rainfall and its estimation is not only important for
India but is equally important for the entire globe
\cite{jang2011importance, tierney2011late, clift2008asian,
fan2013relative, ancy2014prediction, kerhoulas2017influence,
he2006characteristics, lau1984monsoon}.

In India majority of the rain
is received from the month of June to September
(June-July-August-September) and that is why this period is called as
Indian Summer Monsoon Rainfall (ISMR) or the Southwest monsoon rainfall. 
 Cultivated land in India is
majorly benefited by this ISMR \cite{swaminathan1998padma} which makes
this season highly important and ultimately prediction and
estimation of rainfall for this period also becomes equally essential.
India receives nearly 80 percent rainfall during summer monsoon period
\cite{mooley1984fluctuations, naidu2009summer, mooley1997variation} only.
This summer monsoon rainfall fuhrer helps in predicting food grain
production \cite{parthasarathy1988regression} which ultimately
contributes to country's GDP\footnote{https://statisticstimes.com/economy/country/india-gdp-growth-sectorwise.php}. Prediction and estimation of
ISMR started way back from the year 1903 as
people started believing on the importance of this monsoon rainfall  \cite{walker1933seasonal}.

In this work, our area of study is Rajasthan which is the largest state of India and the 60\% of its area falls under the arid category which makes it very environmentally sensitive
\cite{dutta2015evaluating}. Even after being an arid to semi-arid zone, Rajasthan
has observed several floods in the past \cite{goyal2021recommendation,
yadav2021semi, ray2019recent}and also observed
several droughts \cite{karanempirical, bokil2000drought,
goel2006climatic, mundetia2015analysis,
parthasarathy1987droughts}. An early indication of the amount of
monsoon rainfall a particular region is going to receive, can be very
handy in terms of managing the water resource for the entire year. This
early indication can give us an idea about the amount of availability
of water in a particular reservoir. Now this reservoir which will
cater to the needs of demand from people and industry in a particular
area can be regulated and measures can be taken well in advance for
proper water resource management for the monsoon and non-monsoon
period.

There are several indicators on which rainfall depends, such as surface temperature, sea level, distance from sea, distance from mountain ranges etc. In this work, we propose a time series based approach for the prediction of rainfall for the months of June, July, August and September (summer monsoon months). For this we collected Indian Meteorological Department ( IMD hereafter) grided data of 118 years ( from the year 1901 to 2018) and station data of 61 years( from the year 1957 to 2017) from Water Resources Department, Rajasthan (WRD hereafter). In this work we design and analyze advance deep learning models to capture the patterns from this historical time series data for the prediction of Rajasthan summer monsoon rainfall (RSMR hereafter). For this we adapt and improve a model originally proposed by Cheng et al \cite{cheng2016wide} in the field of recommender systems. We name our proposed model as Deep and Wide Monsoon Rainfall Prediction Model (DWMRPM hereafter) and compared with advance deep-learning based models like multi-layer perceptron (MLP), one dimensional convolutional neural network (1D-CNN) based neural networks

\subsection{Related Work}
\label{sec:relatedWork}
In past researchers have applied numerical\cite{ducrocq2002storm} and statistical models\cite{li2010improved, montanari2008estimating} for
rainfall prediction. But with gaining popularity of artificial
intelligence and increasing machine computation power, training  abundant data using machine learning
and deep learning models are becoming the center of
attraction for researchers \cite{ZHANG199835}.  One of the major
reasons of scientists switching from traditional numerical
approaches to artificial intelligence based approaches is that
the statistical and  numerical models fail to capture the dynamic nature of
rainfall \cite{singh2012prediction} whereas neural networks are quiet smart in capturing the
hidden trends and seasonality existing in time series rainfall
data. The numerical and statistical models were used majorly for
two to three decades but these methods lacked forecasting
accuracy \cite{10.2307/24110705} resulting into failure in predicting major rainfall variations \cite{kalsi2004various, sikka2003evaluation}. There are evidences from the
past where these numerical methods failed \cite{gadgil2002forecasting, preethi2011anomalous} to predict the
monsoon rainfall and severe droughts were observed.

Pritpal \cite{singh2018indian} has made an attempt to predict the ISMR using monthly monsoon rainfall values and applied fuzzy sets and artificial neural network (ANN). When the parameters on which the rainfall depends are very high then in order to predict ISMR, \cite{saha2016autoencoder} used auto encoder \cite{ng2011sparse} for reducing the number of parameter and then predicted the ISMR. \cite{saha2021prediction} studied the climatic variables responsible for ISMR and used deep learning feature for monsoon rainfall prediction.This study also shows the monsoon deviation from long period average (LPA) rainfall. Johny et al used an adaptive
Ensemble Model of ANN which was capable of capturing very low and very high rainfall in the Indian state of Kerala \cite{johny2020adaptive}. Dubey et. al \cite{dubey2015artificial} used three artificial neural network based algorithms ( feed-forward back
propagation algorithm, layer recurrent algorithm and feed-forward distributed time delay algorithm) for rainfall prediction over the region of Pondicherry, India. Some amount of monsoon rainfall prediction is done by applying
feed forward neural network \cite{chakraverty2008comparison,
sahai2000all, singh2013indian}. 

Fluctuations in the summer monsoon rainfalls can't be captured efficiently by
traditional linear statistical models \cite{singh2018rainfall,
dash2018indian}. This motivated us to use Deep Learning based
model which are efficient in capturing this non-linearity and
dynamic nature of ISMR. 
As per IMD weather forecasting manual \footnote{https://imdpune.gov.in/Weather/Forecasting\_Mannuals/IMD\_IV-13.pdf}, Indian rainfall is very well known for its variability in space and time. There is hardly any seasonal distribution of rainfall over entire India. At two different station locations which are a few miles apart, if we consider one day rainfall, we may observe that one station experiencing heavy rainfall whereas the other station may go completely dry. This kind of variation is not only found in monsoon rainfall period(June to September) but also during post monsoon period (October to December) as well.   

A good amount amount of work has been done in the field of ISMR
as presented above but at present to the best of our knowledge, no work is done in the field
of RSMR prediction, which attracted
the authors of this paper to explore this untouched area. An attempt to predict the agricultural drought index in Rajasthan is done by Dutta et al \cite{dutta2013predicting} using standardized precipitation index.
In this proposed work, an extensive
study is done in predicting RSMR for the first time. The good
thing about Rajasthan is the strong Rain Gauge network from
IMD, Water Resource
Department, Rajasthan and the Revenue Department which has
resulted into the abundant supply of rainfall data for analysis
and prediction.

Research work done in the field of Rainfall Prediction and
Estimation for the state of Rajasthan is very less.
Vikas et al \cite{bajpai2020prediction} used the historical time-series data
for daily rainfall prediction. However worked in analyzing the trends of rainfall in the state of
Rajasthan\cite{ pingale2014spatial, yadav2018analysis}. \cite{bryson1981holocene} made an effort to present the holocene variations of monsoon rainfall in Rajasthan. \cite{meena2019trends} made an attempt to explore the spatial and temporal differences to identify trends in monthly, seasonal and annual rainfall over the Rajasthan region. They observed the prevailing homogeneity of rainfall at various stations in the state.
 In another work \cite{singh2012probability} authors tried to estimate the one day maximum
rainfall in Jhalrapatan, a city in the state of Rajasthan.
Authors have done the probability analysis for this purpose.
\cite{lal2020study} studied the rainfall pattern in Chaksu,
Rajasthan.

Our objective is to predict the RSMR which starts in the month of June and ends in the month of September. In this work we propose a time series based prediction model which depends on the fundamental of present and future time series data dependency on past time series data \cite{singh2016applications}. We adapt and improvise wide and deep learning model originally proposed by Cheng et al \cite{cheng2016wide} for recommendations. Many authors h
ave used this concept in different domains like regression analysis \cite{kim2020wide}, quality prediction \cite{ren2020wide}, rainfall prediction \cite{bajpai2020prediction} etc. Wide networks are used for memorization and deep networks are used for generalization. In this work we propose a Deep and Wide Monsoon Rainfall Prediction Model (hereafter DWMRPM) to predict monsoon rainfall prediction in the Indian state of Rajasthan.   

The rest of the paper is organized as follows. Section 2 explains the proposed model for summer monsoon rainfall in Rajasthan. Details of experimental evaluations, model training, results of rainfall prediction and comparison with other deep learning approaches is given in section 3. Finally we conclude the paper in Section 4 and provide avenues for future work.

\subsection{Major Contributions}
\begin{enumerate}
	\item In this work, we propose a novel architecture based on deep and wide neural network for the purpose of summer monsoon rainfall using historical time-series data. The model efficiently captures the dynamic nature of monsoon rainfall and works well in its  prediction. To the best of our knowledge, we are the first who have tried to solve this challenging problem. 

	\item We compare our work with various advanced deep learning algorithms for sequence prediction on two different types of datasets and have obtained very promising results.

	\item The algorithms we designed has the generalization ability and can be used to predict summer monsoon rainfall for atmospherically different regions of Rajasthan.

\end{enumerate}

\section{Deep \& Wide Monsoon Rainfall Prediction Model (DWMRPM)}
\label{sec:DWMRPM}

This section first provides a brief overview of the proposed approach
and subsequently explain various steps involved in the prediction of
Rajasthan summer monsoon rainfall (RSMR, hereafter).

%
%
%
%
\subsection{Overview}
\label{subsec:overview}

In this work we address the problem of summer monsoon rainfall in
Rajasthan, which is the largest state of India and is located in the
North-Western part of the country. Rajasthan has very distinct physiographic
characteristics.
On one side it has India's biggest desert area, called The Thar
Dessert and on the other side this state
has Eastern Plains and the ranges of Aravalli Hills \cite{enzel1999high}. These ranges are
in the direction of South-west monsoon, which is responsible for
rainfall in the region \cite{roy2002geology}.
Atmospherically Rajasthan is divided into four zones: North West Desert Region, Central Aravalli Hill Region, 
Eastern Plains and South Eastern Plateau Region \cite{Upadhyaya14}. 
Details of the districts, which come under the respective zones are given below:
\begin{itemize}[]
	\item[] \textbf{North-West Desert Region:} Jaisalmer, Jodhpur, Hanumangarh, Shriganganagar, Barmer, Churu, Nagaur, Pali, Sikar, Bikaner and Jhunjhunu 
	\item[] \textbf{Central Aravalli Hill Region:} Udaipur, Dungarpur, Sirohi, Jalore, Pali, Banswara, Bhilwara, Chittorgarh, Rajsamand and Ajmer
	\item[] \textbf{Eastern Plains:} Alwar, Bharatpur, Tonk, Sawai Madhopur, Karauli, Jaipur, Dausa and Dhoulpur
	\item[] \textbf{South-Eastern Plateau Region:} Kota, Bundi, Jhalawar and Baran 
\end{itemize}

All these zones have different atmospheric and climatic conditions. The problem of predicting summer monsoon rainfall in Rajasthan is different
from the prediction of Indian summer monsoon rainfall (ISMR,
hereafter). Most
of the time-series-based methods for predicting ISMR consider average
monthly rainfall values by taking weighted average of the 306 well
distributed rain-gauge stations in the non-hilly areas of Indian
sub-continent \cite{dash2018indian, singh2018rainfall,
singh2013indian, sahai2000all}. 
Rajasthan being a dry state lies in arid and semi-arid
zones and characterized by low and uneven rainfall
\cite{kulshreshtha2013majestic}, therefore, a dedicated system is required which can predict monsoon rainfall for different geographical regions separately.
We use historical monthly rainfall data from two different sources to train
and analyze the performance of our model in prediction of Rajasthan
Summer Monsoon Rainfall. Details on the datasets are given in
Section~\ref{subsec:Dataset}

For ISMR researchers used monthly rainfall values of June to September across all the years \cite{dash2018indian} or just have captured the dependency of months of a single year \cite{singh2018indian}
In order to avoid loss of any information, we are using rainfall
values of all the months of previous years for the prediction of
rainfall for the months of June, July, August and September.
For example in order to predict rainfall for the month of June 2019,
we use rainfall values of all the months from May 2000 to May 2019.

\begin{figure}[htb]
	\centering
	\includegraphics[scale=0.56]{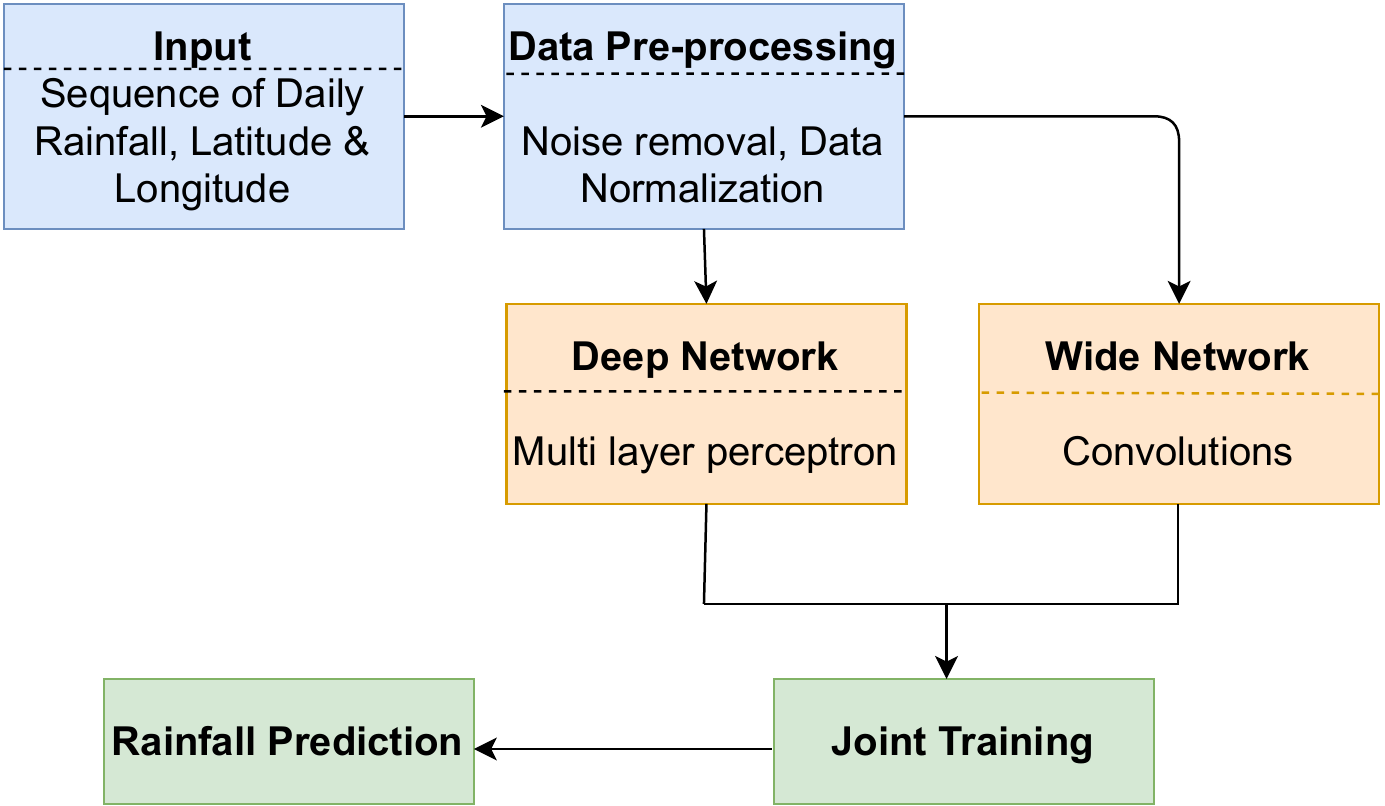}
	\caption{Overview of DWMRPM: The model takes sequence of monthly rainfall 
	intensities and geographical parameters, namely
	latitude and longitude as input. After initial
	pre-processing, input is fed to a deep network, which is a
	multi-layer perceptron, and to a wide network, which is a
	convolutional network. The model is jointly trained considering the
	activation weights from both deep and wide networks simultaneously. }
	\label{fig:Overview}
\end{figure}

In this work, we propose a deep and wide monsoon rainfall prediction
model (DWMRPM)
for the prediction of the total monthly rainfall intensity for the
summer monsoons months of Rajasthan. The wide network is used to
extract low-dimensional features. Here, instead of using a sequence of
monthly rainfall values directly,
we are using features obtained after applying a convolutional
layer, as
it is very effective in learning spatial dependencies in and between
the series of data \cite{Van20}.
High-dimensional features, on the other hand, are derived using
Multi-layer perceptron (MLP) \cite{pal1992multilayer} in which a 
sequence of rainfall intensity values are passed on to a deep network.
In order to incorporate a geographical generalization ability
in the model, so that a single model can be used to make
rainfall predictions in different geographical conditions,
information of geographical parameters (latitude and longitude)
is included at the time of training.
The operational steps involved in the development of our proposed
DWMRPM for the prediction of rainfall are shown in
Figure~\ref{fig:Overview}.

To evaluate the performance of the
proposed method, we use two standard statistical metrics, namely
mean absolute error (MAE) and root mean square error (RMSE). We compare
our results with the advance deep learning models
like MLP and one dimensional convolutional neural networks
(1-DCNN) which are very
popular for sequence based predictions.


\subsection{Dataset description and pre-processing}
\label{subsec:Dataset}
\begin{figure*}[!htb]
\centering
	\subfloat[]{\includegraphics[scale=0.5]{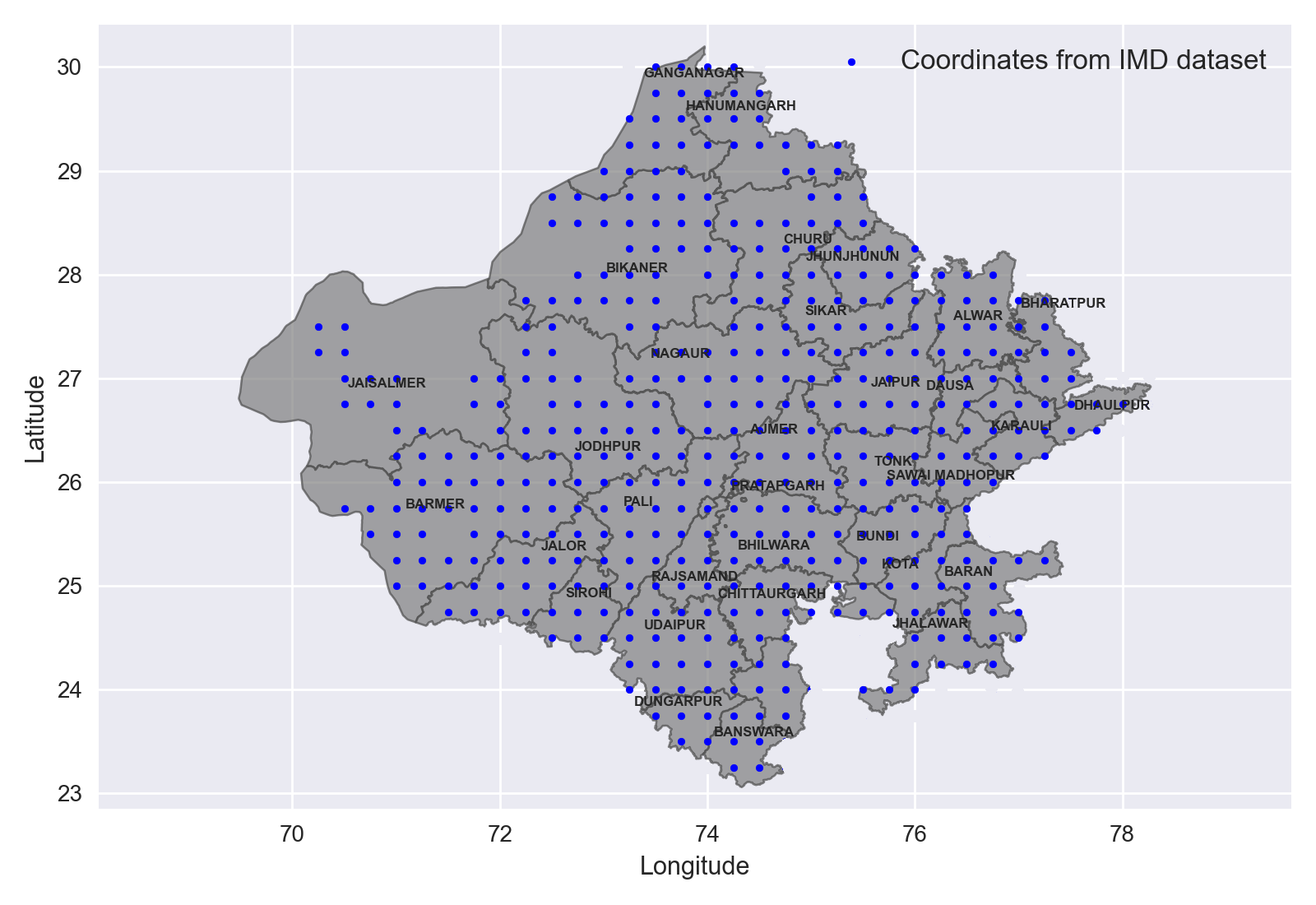}
	\label{fig:rajasthanMapIMD}} \\
	\subfloat[]{\includegraphics[scale=0.5]{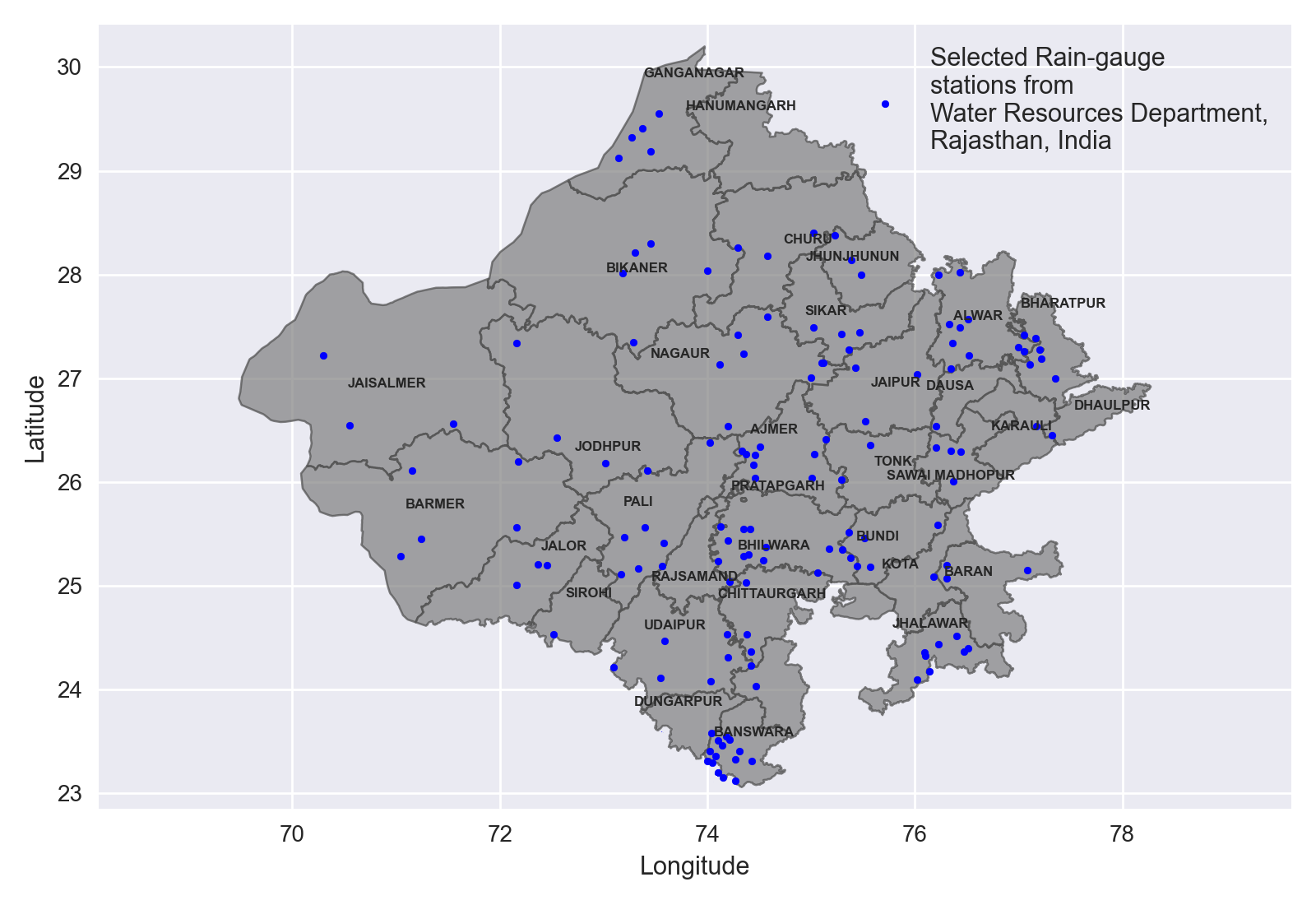}
	\label{fig:rajasthanMapWRD}} 

	\caption{Map of Rajasthan showing distribution of (a) selected 484
	coordinates from IMD gridded dataset at high
	 spatial resolution of $.25^\circ \times .25^\circ$, (b) selected
	 158 rain-gauge stations from Water Resources Department,
	 Rajasthan. The rainfall data obtained from these coordinates and
	 rain-gauge stations are used for the prediction of Rajasthan
	 summer monsoon rainfall (RSMR)}
\label{fig:rajasthanMap}
\end{figure*}

In this work we have used Water Resources Department dataset and Indian Meteorological Department (IMD)
gridded rainfall data with a high spatial
resolution of $0.25^\circ \times 0.25^\circ$ \cite{pai2014development}. 
From IMD data set, we selected the
rainfall data of the Rajasthan meteorological sub-division
ranging from $23^\circ3.5'$N to $30^\circ14'$N latitude and
$69^\circ27'$E to $78^\circ19'$E longitude, for the period of 118
years from the year 1901 to 2018. It gave the rainfall data for
1008 rain-gauge stations.
We have also collected the rainfall data from Rajasthan's water resources
department, for more than 500 rain-gauge stations, over a period
of 61 years (from the year 1957 to 2017).
The datasets were noisy in terms of negative and missing
values. After initial level data pre-processing and cleansing steps, we
selected 484 co-ordinates from IMD dataset of High Spatial Resolution of (0.25X0.25 degree) and 158
stations from Rajasthan's water resources data for our analysis.
The distribution of the selected stations from water resources
data, over 33 districts are depicted on
the map of Rajasthan in Figure~\ref{fig:rajasthanMap}

In this paper, authors have made use of both the Station data (data collected from various Rain Gauge Stations in Rajasthan) and the Gridded data \cite{pai2014development}. The idea behind using both the data sets is that when only the station data is used, for experimentation, one uses the data for single point of scale whereas when the gridded data is used the application of different meteorological data for a region is applied depending upon the resolution. In large catchment areas where less number of Rain Gauges are installed, modeling may not be that much accurate, on the other hand gridded data is more continuous and may prove better than single point estimates. 
Gridded data contains the data from stations or satellites (in our case, it's rain gauge station data) which undergoes interpolation over a grid. This interpolation needs careful analysis for biases and outliers \cite{rajeevan2005development, pai2014development}. Station data on the other hand is unbiased single point data. 
For our study, we have used quality controlled data sets from both the categories.
If someone has enough single point station in the region under study,
then the station data can be easily utilized but since the rain gauge
distribution is not uniform ( as shown in Raj\_WRD), specially in the dessert areas where the rain gauge station installation density is very low, combination of both the data set seems to be optimal. Another advantage of using gridded data is that it acts as a source of replacement to the data missing from the records of rain gauge stations\cite{meher2019gridded}. Any area or zone where the observed station data (point data) is comparatively less, interpolated gridded data can work as a potential alternative means \cite{bandyopadhyay2018comparison}.

These datasets contained daily rainfall values from which we
calculated monthly rainfall values for January to December. In
order to provide rainfall pattern in Rajasthan, mean rainfall
values for each month and the monsoon season (combined rainfall
of June, July, August and September) from the year 1901-2018 for a
randomly picked rain-gauge station is shown in Table~\ref{tab: StatisticalSummary}. We also
provide the minimum and maximum rainfall for each month and
monsoon season, over the
duration of 118 years. It can be
observed that the significant amount of annual rainfall occurs in
the monsoon months and the in the remaining months, the aggregate 
rainfall is very less.

\begin{table}[htb]
	\caption{Statistical summary of monthly data for the IMD dataset and
	the dataset from the water resource department of Rajasthan (WRD). Mean, maximum and
	minimum rainfall fall values of IMD are shown for a randomly picked
	coordinate at $26^\circ0'$N and $74^\circ5'$E for a period of 118 years
	from 1901 to 2018. The rainfall statistics for WRD dataset for a
	randomly picked rain-gauge station, situated at $26^\circ04'$N and
	$75^\circ01'$E is shown for a period
	of 61 years from the year 1957 to 2017.}	
	\label{tab: StatisticalSummary}
\centering
\hspace*{-0.5cm}
	\begin{tabular}{lllllll}
\hline
		\multirow{2}{*}{Month} & \multicolumn{2}{c}{Mean (mm)} & \multicolumn{2}{c}{Maximum (mm)} & \multicolumn{2}{c}{Minimum (mm)} \\
		\cline{2-7}
		& IMD & WRD & IMD & WRD & IMD & WRD \\
		\hline

\hline
Jan & 4.24 & 2.20 		& 63.66 & 53.00 & 0 & 0 \\
Feb & 4.22 & 2.40 		& 65.56  &54.00 & 0 & 0 \\
Mar & 3.81 & 2.11 		& 64.44  &74.00 & 0 & 0\\
Apr & 2.91 & 3.78 		& 43.03  &56.00 & 0 & 0\\
May & 9.14 & 5.72 		& 90.75  &87.00 & 0 & 0\\
Jun & 49.67 & 41.92		& 246.2  &227.00 & 0 & 0\\
Jul & 159.45 & 150.54  	& 523.5  &476.00 & 13.11& 10 \\
Aug & 160.86 & 166.05 	& 441  & 905.0& 3.41& 40.8 \\
Sep & 65.38 & 63.83 		& 305.8  &402.00 & 0& 0\\
Oct & 9.84 & 7.31 		& 154.4  & 132.00& 0& 0 \\
Nov & 1.72 & 4.02 		& 25.67  &160.00 & 0& 0\\
Dec & 2.31 & 1.04 		& 46.53  & 30.00& 0& 0\\
Overall Accuracy &  868.63 & 460.05 & 1080.90  & 937.0 & 91.17 & 102.1 \\
\hline

\end{tabular}
\end{table}

We have considered time-series values of monthly rainfall and
geographical parameters like latitude and longitude for the
prediction of rainfall during the monsoon months in different
regions of Rajasthan. The rainfall intensity values ranges from 0 mm to
more than 800 mm while coordinate values of latitude and
longitude lies between
$23^\circ3.5'$N to $30^\circ14'$N and $69^\circ27'$ to
$78^\circ19'$E, respectively. Since the data is of different
dimensions and dimensional units, therefore we normalize the
data to make it dimensionally uniform. When the magnitude of
different parameters in a dataset is different, the parameters
with higher values suppresses the role of the parameters with lower
values in model training. To handle this issue, we use the
min-max normalization method to convert all rainfall intensity
values to number between 0 and 100 (latitude and longitude values
are already in this range). The mathematical representation of the min-max normalization
method is as follows:

\begin{equation*}
	I^{*} = \frac{I - I_{min}}{I_{max} - I_{min}} \times 100
\end{equation*}

where, $I^{*}$ is the normalized value of the
monthly rainfall intensity value, $I$ represents a value in the
original dataset,
$I_{max}$ and $I_{min}$ are the maximum and minimum intensity values, respectively.
Normalization can also help in 
	improving the learning capability of the model
	and in reducing the computational complexity
	\cite{shanker1996effect}.

\subsection{Model Description}
We use deep and wide neural network-based architecture \cite{bajpai2020prediction} 
for the purpose of summer monsoon rainfall prediction in the Indian state of Rajasthan. The
following paragraphs explains the major components of the model.

\subsubsection{The Wide Component: Convolutions}

The wide component is used to memorize
certain combinations of monthly rainfall events, which is beyond the
capabilities of the deep model. It is a generalized linear
model of type $y = \mathrm{\textbf{w}^{T}\textbf{x}} + b$.
In the model proposed by Cheng \textit{et al} \cite{Cheng16},
 cross-product feature
transformations were used as the wide component. 
In this work we use convolutional network as wide component.
The basic components of a general CNN consists of 2 types of
layers, namely convolutional layer and pooling layer
\cite{gu2018recent}. The convolutional layer is composed of
several convolutional kernels, which capture and learn the correlation of
spatial features by computing different feature maps. The output
of one dimensional convolutional layer with input size $N_{l}$
is:
\begin{equation*}
	a^{(l+1)}_{k} = b^{(l+1)}_{k} + \sum_{i=1}^{N_{l}} conv1D(w^{l}_{i,k}, a^{l}_{i})
\end{equation*}
where, $l$ is the layer number, $w^{l}_{i,k}$ is the kernel from
the $i^{th}$ neuron at layer $l-1$ to the $k^{th}$ neuron at
layer $l$, $a^{(l)}$, $b^{(l)}$ activations, bias at $l^{th}$ layer.

\begin{figure}[htb]
	\centering
	\includegraphics[scale=0.6]{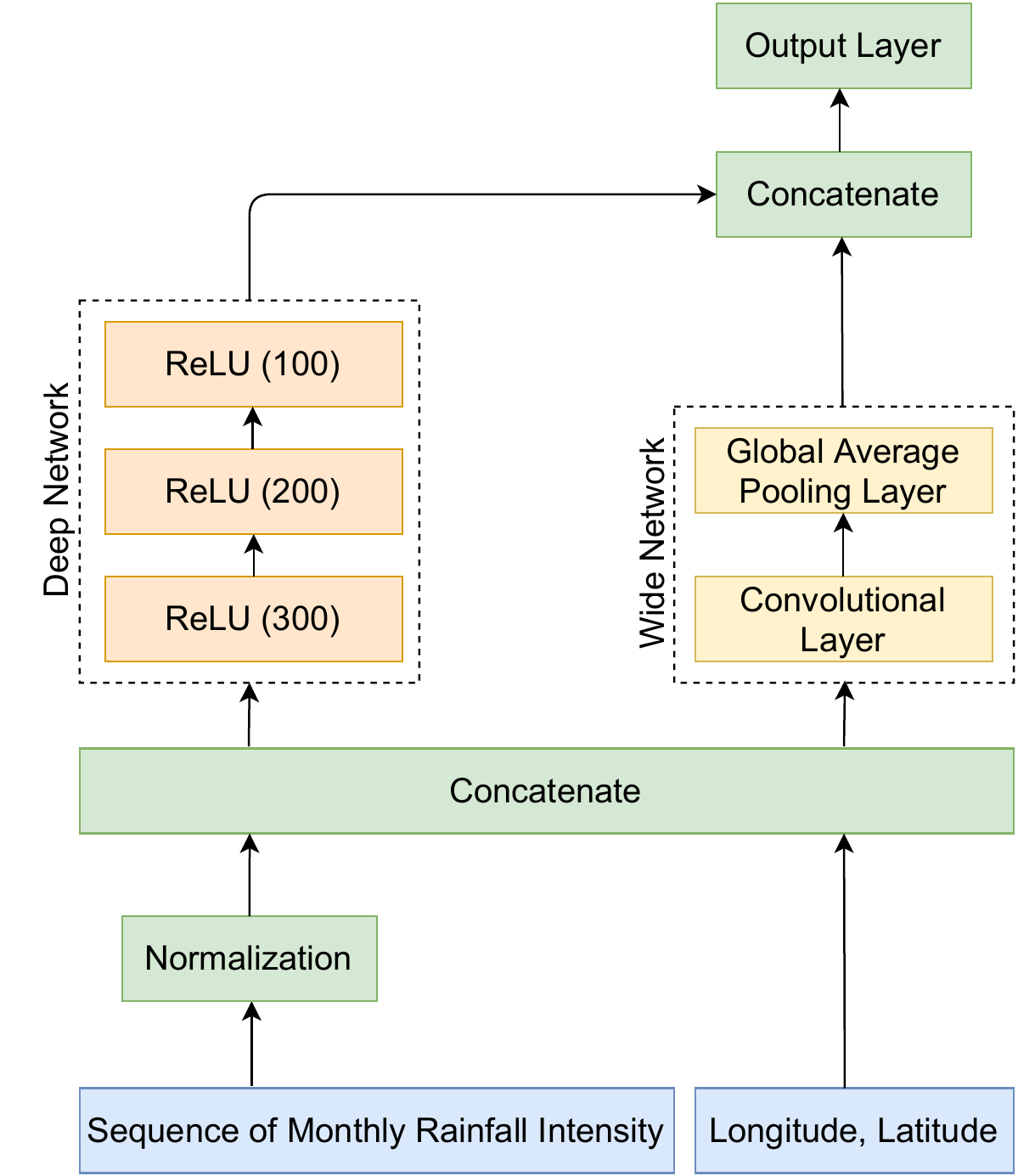}
	\caption{Selected architecture of DWMRPM for prediction of Rajasthan Summer Monsoon Rainfall. There are two major components: 1.The Deep
	component consists of mainly an input layer and 3 ReLU
	layers. 2. The wide component consists of a convolutional
	layer followed by a global average pooling layer.A sequence of monthly rainfall intensity values after normalization and values are fed to deep and wide components separately }
	\label{fig:Architecture}
\end{figure}

Convolutional layer is followed by a pooling layer that is used
to realize shift invariance by reducing the resolution of the
feature maps.
As demonstrated by 
\cite{Van20}, 1D
CNN performs well in regression type
of problems and can learn to find the correlation in between the
series very efficiently. Therefore, instead of using raw features
in the wide part of the network, we use a convolutional layer
to capture such combinations. 
In addition to this, to make our model more
generalized with respect to different atmospheric conditions, we
are using geographical parameters namely, longitude and
latitude while designing and developing our model (Figure~\ref{fig:Architecture}).

\subsubsection{The Deep Component: Multi-layer Perceptron}
The deep component is a feed-forward neural network, specifically
a multi-layer perceptron, as shown in
Figure~\ref{fig:Architecture}.
Sequence of monthly rainfall intensity values are given
as input, which are then fed into hidden layers of a neural
network in the forward pass. Typically, each hidden layer
computes:
\begin{equation*}
	a^{(l+1)} = f(w^{l}a^{l} + b^{l})
\end{equation*}
where, $l$ is the layer number and $f$ is the activation
function, rectified linear units (ReLUs) in our case, $a^{(l)}$,
$b^{(l)}$, and $w^{(l)}$ are the activations, bias and model
weights at $l^{th}$ layer.

\subsubsection{Joint training of the model}
\label{subsubsec:jointTraining}
The model is trained using the joint training approach that
optimizes all parameters simultaneously by taking into
account the output of the deep and wide components and their weighted sum. 
It helps in providing
an overall prediction, which is based on aforementioned
components, also depicted in Figure~\ref{fig:Architecture}. 
\begin{equation*}
	y_{DWMRPM} = \textbf{k}_{cn}\textbf{h}_{cn} + \textbf{k}_{d}\textbf{h}_{d}
\end{equation*}
where, $y_{DWMRPM}$ is the prediction, $\textbf{h}_{cn}, \textbf{h}_{d}$ are the output vectors of two
sub-models namely wide-convolutional model and deep model respectively, and $\textbf{k}_{cn}, \textbf{k}_{d}$ are their respective weight
vectors to be trained.

\section{Experimental evaluations}
\label{sec:experimentalEvaluations}
\subsection{Implementation details}
All the experimental programs are coded using Keras \cite{chollet2015keras}
API of TensorFlow framework \cite{abadi2016tensorflow, gulli2017deep}.
The hardware setup includes computer processor from Intel with i7-8750H configuration supported by 32GB RAM.
The upcoming sections and subsections describe the designing and implementation setup of proposed approach and baseline approaches followed by results obtained.

For prediction of monthly rainfall of monsoon season, we consider different training windows of lengths ranging from 2 years to 10 years. We found that 9 years training window gives most accurate prediction results for the monsoon months of June, July August and September.

\subsubsection{Training, validation and test sets}
\label{subsubsec:trainingTestSets}

We use two type of datasets, one from the Indian
Meteorological Department (IMD) and the other from the Water Resource
Department (WRD).
In case of WRD, monthly rainfall values from the year 1957 to 1986 are
considered for the purpose of training. Validation of the model is
done on the dataset considering the years  starting from 1987 to 1997
and finally we test the model on the dataset containing monthly
rainfall values in the interval of
the year 1998 to 2017. In case of IMD dataset, training is done by
considering values from the year 1901 to 1980 and validation is done
from the year 1981 to 1995 and finally testing is done on the rainfall
intensity values from the year
1996 to 2018. 

\subsubsection{Evaluation metrics}
\label{subsubsec:evaluation}
As shown by \cite{Glorot10} and \cite{He15}, to evaluate the overall
accuracy of predictions, we use root mean square error (RMSE) and mean
absolute error (MAE) as the basic evaluation metrics. Low value of
RMSE and MAE means better prediction accuracy of the model.

\begin{align}
	\label{eq:evaluationMetrics}
	RMSE &= \sqrt{\frac{1}{N}\sum^{N}_{i=1}\bigg{(}y_{i}-\overline{y}_{i}\bigg{)}^{2}} \\
	MAE &= \frac{1}{N}\sum^{N}_{i=1}\Big{|} y_{i}-\overline{y}_{i} \Big{|}
\end{align}

where, N represents the number of samples, $y_{i}$ is the
actual rainfall of the $i{th}$ sample and
$\overline{y}_{i}$ is the corresponding prediction.

\subsubsection{Model Training}
\label{subsubsec:modelTraining}

We optimize various hyper parameters like the batch size, number of
hidden layers, number of neuron and the
dropout rates using trial-and-error
method. The network configuration of DWMRPM used in our
experiments is shown in Figure~\ref{fig:Architecture}.  The input to the model is the normalized sequence of monthly rainfall intensity values and actual coordinate values (latitude and longitude). The deep part is a Multi-layer perceptron with an input layer;
3 hidden layers
containing 300, 200 and 100 neural units with ReLU as the
activation function; and finally a dense output layer.In order to prevent over-fitting of the
model, dropout layers \cite{srivastava2014dropout}
with dropout rate 0.3 are added after each hidden
layer. The wide part contains a convolutional layer with 100 filters,
each of size 1 x 5, followed by a global
average pooling layer. The outputs of both the wide and deep networks are concatenated and the model is trained
using the joint-training approach, as explained in
Section~\ref{subsubsec:jointTraining}.
We use Adam optimizer \cite{kingma2014adam} for
training with Mean Square Error (MSE)  as loss
function, which is calculated as follows:
\begin{equation*}
		MSE = \frac{1}{N}\sum^{N}_{i=1}\bigg{(}y_{i}-\overline{y}_{i}\bigg{)}^{2}
\end{equation*}
Here, N represents the number of samples, $y_{i}$ is the
actual rainfall of the $i{th}$ sample and
$\overline{y}_{i}$ is the corresponding prediction.
The goal of the model is to find optimized parameters
that minimizes MSE
\begin{equation*}
	\underset{\theta}min MSE(\theta)
\end{equation*}
where, $\theta$ is the total
number of trainable parameters.
Weights of the network are initialized using He
initialization\cite{he2015delving}.
Model is trained for 200 epochs with batch size equals to 8.

\begin{figure*}[!ht]
\centering
	\subfloat[]{\includegraphics[scale=0.5]{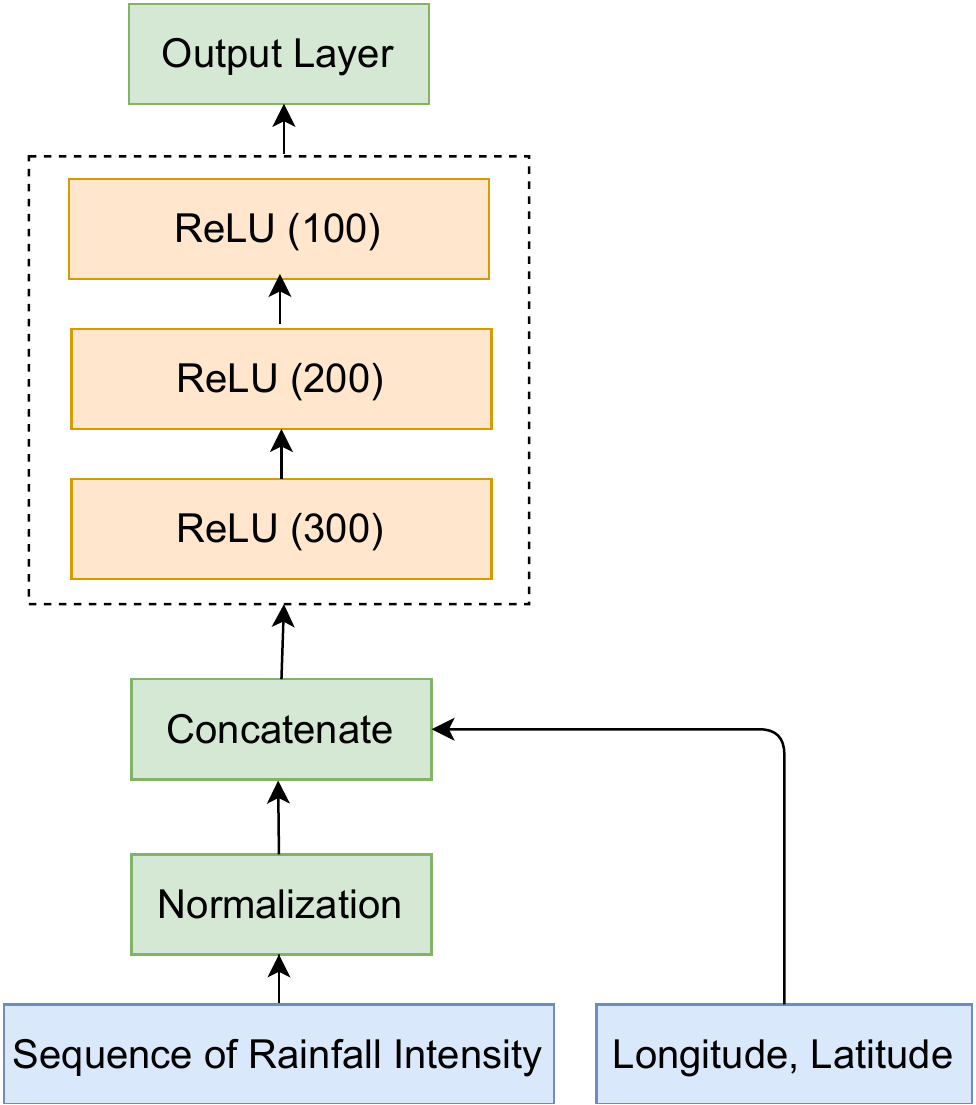}
\label{fig:MLPArch}} \hspace{1mm}
\subfloat[]{\includegraphics[scale=0.5]    {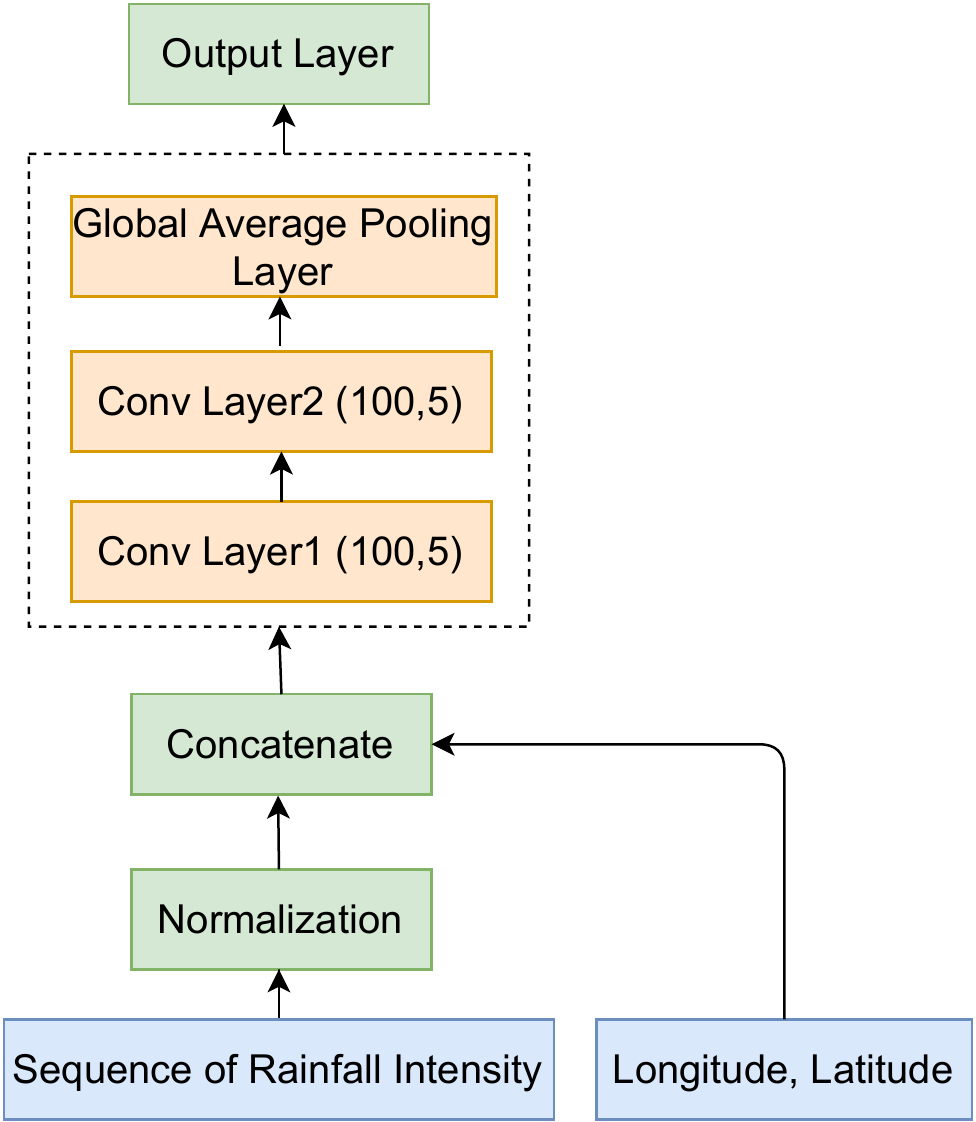}
\hspace*{-3cm}
\label{fig:CNNArch}} 
\caption{Architecture of the baseline approaches, selected after
	experimentation with various hyper-parameters.  (a) Network
	configuration of multi-layer perceptron, and (b) Network configuration of CNN.}
\label{fig:BaselineArch}
\end{figure*}

\subsubsection{Baseline approaches}
\label{subsubsec:baselineApproaches}
In order to establish the competence of our proposed approach, we have
compared the results obtained from the proposed DWMRPM with the
results of two advance deep learning approaches: MLP and 1-DCNN. These
approaches are working well but not at par with our proposed approach. 
We have used the same sets of both the data sets for all the
approaches in order to avoid any discrepancies that may arise by using
different set of datasets for training and testing the models. 
The network architecture of the baseline approaches, which is
selected (after experimenting with various hyper-parameters)
for the comparative analysis with the proposed method is
explained in the subsequent paragraphs.
In all these approaches, we use Adam optimizer for training
and MSE as loss function. Input sequence length is 108 (9 years).
(Details in Figure~\ref{fig:BaselineArch})

\begin{itemize}[]

	\item[] \textbf{Multi-layer perceptron (MLP):} 
		The network architecture for MLP is shown in
		Figure~\ref{fig:MLPArch}.Sequence of rainfall is normalized and concatenated with latitude and longitude. It contains 3 hidden ReLU layers with 300, 200 and 100 units of neurons respectively.

	\item[] \textbf{Convolutional Neural Network (CNN):} The network
		architecture selected for CNN is given in
		Figure~\ref{fig:CNNArch}.
		Sequence of rainfall is normalized and concatenated with latitude and longitude. The setup has two convolutional layers with 100 filter size of 1x5 each followed by Global Average Pooling layer.
\end{itemize}

\subsection{Results and discussion}
\label{sec:results}
In the following subsections, we present the results of
experimental analysis and comparison of the proposed method with
the baseline approaches described in Section~\ref{subsubsec:baselineApproaches}.

\subsubsection{Forecasting accuracy of DWMRPM}

As mentioned in Section~\ref{subsec:overview}, Rajasthan is divided
into four atmospheric zones, each of which having huge difference in
their climatic and physiographic properties. To evaluate the
effectiveness and accuracy of the proposed model, we apply it on each
zone separately. The prediction results on one of the randomly picked
rain-gauge stations from each zone is given in
Table~\ref{tab:zonePred} and graphical representation is shown in
Figure~\ref{fig:Results}. Here we have used station data from WRD. IMD
gridded data is not used in this case because the dataset is generated by interpolation which may have biases and outliers \cite{rajeevan2005development, pai2014development}.

\subsubsection{Generalization ability of DWMRPM}
\label{subsubsec:generalization}
\begin{figure*}[!ht]
\centering
	 \hspace*{-1.5cm}
	\subfloat[]{\includegraphics[scale=0.50]{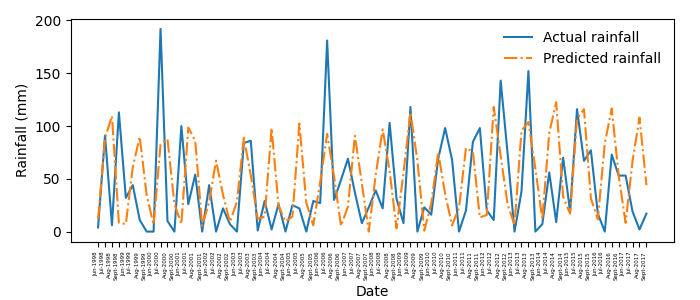}
	\label{fig:Z1}} 
	\subfloat[]{\includegraphics[scale=0.50]{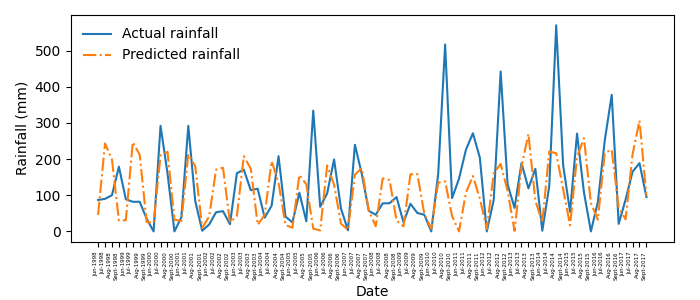}
	\label{fig:Z2}}	\\
	 \hspace*{-1.5cm}
	\subfloat[]{\includegraphics[scale=0.50]{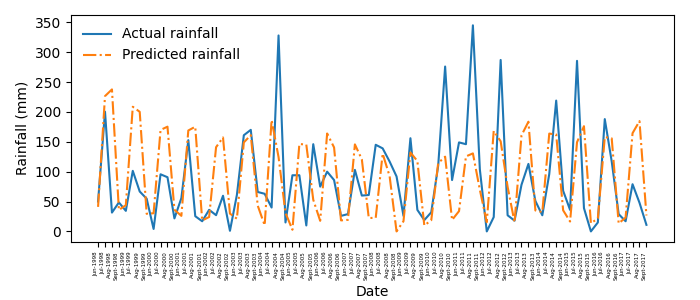}
	\label{fig:Z3}} 
	\subfloat[]{\includegraphics[scale=0.50]{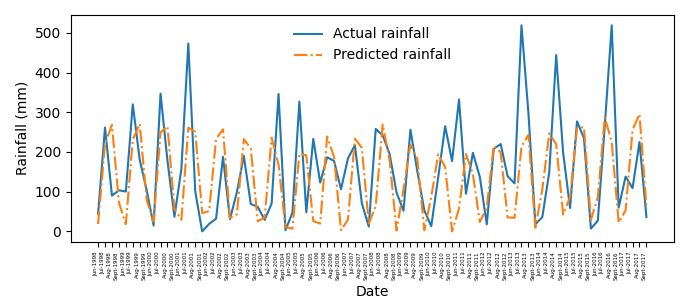}
	\label{fig:Z4}}
\caption{Prediction results of DWMRPM for four rain-gauge
	stations, each picked from a different atmospheric zone
	(Section~\ref{subsubsec:generalization}). Results are from
	year 2016 to 2017. (a) Prediction
	results of rain-gauge station situated at 29$^\circ$12'N,
	73$^\circ$14'E in North-West dessert region, (b) Prediction
	results of rain-gauge station situated at 26$^\circ$04'N,
	74$^\circ$46'E in Central Aravalli hill region, (c) Prediction
	results of rain-gauge station situated at 26$^\circ$41'N,
	75$^\circ$14'E in Eastern plains region, and (d) Prediction
	results of rain-gauge station situated at 25$^\circ$18'N,
	75$^\circ$57'E in South-Eastern plateau region.}
\label{fig:Results}
\end{figure*}

	In order to verify generalization
	ability of our model, we use it for monsoon rainfall prediction in
	each zone separately.  
	The prediction results for each zone,
	on the
	basis of two evaluation criteria i.e., MAE and RMSE
	(Section~\ref{subsubsec:evaluation})  on WRD
	dataset
	are shown in Table~\ref{tab:zonePred} and Figure~\ref{fig:Results}. 

	\begin{table}
	\caption{Prediction results of DWMRPM on the four rain-gauge
	stations from
	WRD dataset. Each station is randomly picked from four different
	atmospheric zones.}
	\label{tab:zonePred}
	\hspace*{-1.2cm}
	\begin{tabular}{p{2.5cm}p{1cm}p{1cm}p{0.7cm}p{0.7cm}p{0.7cm}p{0.7cm}p{0.7cm}p{0.7cm}p{0.7cm}p{0.7cm}}
		\hline
		\multirow{2}{*}{Zone Name} & \multirow{2}{*}{Latitude} &  \multirow{2}{*}{Longitude} & \multicolumn{2}{c}{June} & \multicolumn{2}{c}{July} & \multicolumn{2}{c}{August} & \multicolumn{2}{c}{September} \\
		\cline{4-11}
		&&& MAE & RMSE & MAE & RMSE & MAE & RMSE & MAE & RMSE \\
		\hline
		North-West Desert& 29$^\circ$12'N & 73$^\circ$14'E & 2.4164 & 2.7660  & 5.1061 & 5.8879 & 4.8080 & 5.6060   & 2.8041 & 3.3493 \\  
		Central Aravalli Hill Region & 26$^\circ$04'N &
		74$^\circ$46'E & 3.1414 & 3.7925  & 7.7769  & 9.2311 & 11.5339 & 13.3020 & 4.7901 & 6.8593 \\
		Eastern Plain& 26$^\circ$41'N & 75$^\circ$14'E & 3.1223 & 3.7269  & 6.5146  & 7.8600 & 9.5284  & 10.7145 & 2.8730 & 3.4377 \\
		South-Eastern Plateau Region & 25$^\circ$18'N &
		75$^\circ$57'E & 6.2528 & 7.5165  & 8.3010  & 10.3782 & 9.0789 & 10.7850 & 5.0400 & 5.8975 \\
\hline
\end{tabular}
\end{table}

It can be observed that a single model is working well in rainfall forecasting for different geographical conditions ranging from plains and plateaus to desserts and hills.  	

\subsubsection{Comparison with baseline approaches}
\begin{figure*}[!ht]
\centering
	\subfloat[]{\includegraphics[scale=0.50]{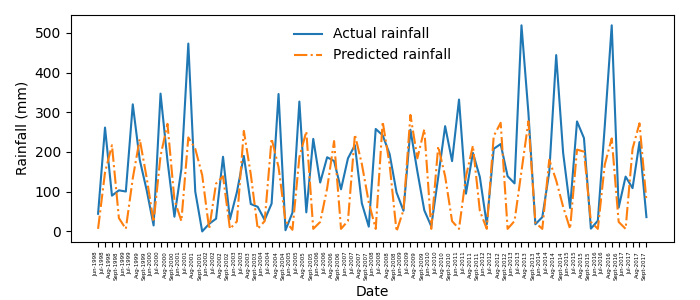}
	\label{fig:A1}}\\ 
	\subfloat[]{\includegraphics[scale=0.50]{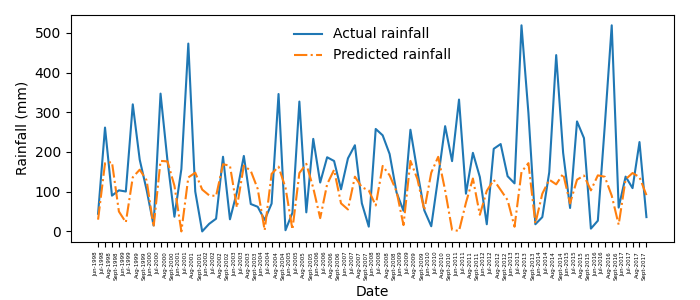}
	\label{fig:A2}}\\
	\subfloat[]{\includegraphics[scale=0.50]{DWRPM_WRD_2518_7557.png}
	\label{fig:A3}} 
\caption{Comparison of DWMRPM and two deep-learning approaches on WRD
	dataset for a randomly selected rain-gauge station situated
	at 25$^\circ$18'N, 75$^\circ$57'E. The results are for the months
	of June, July,
	August and September of 20 years (June 1998 to September 2017)
	(a) Prediction
	results of MLP, (b)
	Prediction results of one dimensional CNN and, (d) Prediction results of the
	proposed DWMRPM.}
\label{fig:ComparisonWRD}
\end{figure*}

\begin{figure*}[!ht]
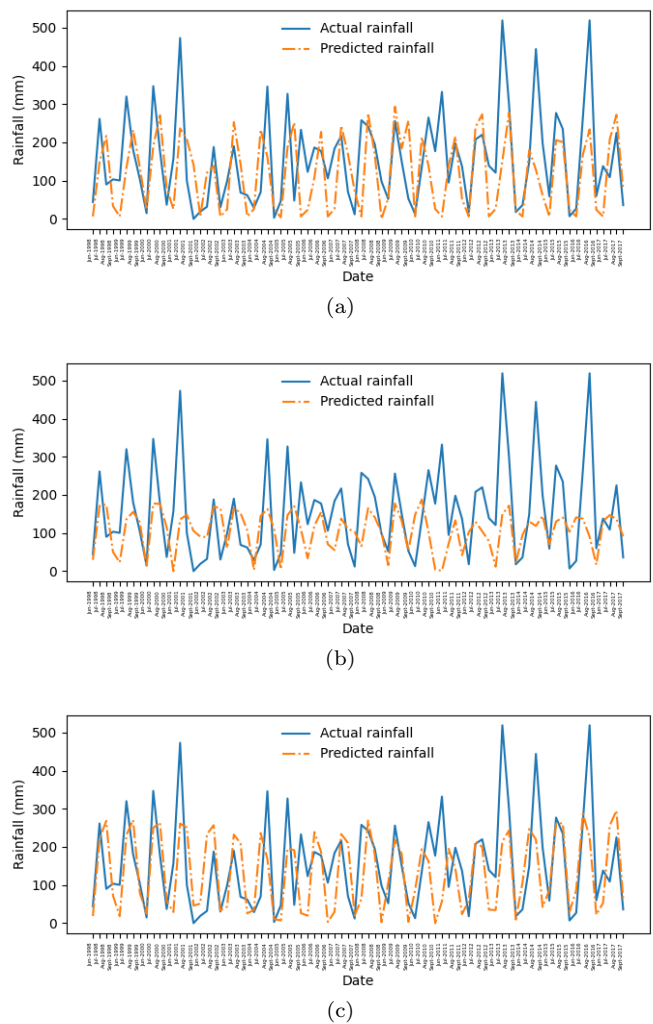

\centering
	\subfloat[]{\includegraphics[scale=0.50]{MLP_WRD_2518_7557.png}
	\label{fig:A1}}\\ 
	\subfloat[]{\includegraphics[scale=0.50]{CNN_WRD_2518_7557.png}
	\label{fig:A2}}\\
	\subfloat[]{\includegraphics[scale=0.50]{DWRPM_WRD_2518_7557.png}
	\label{fig:A3}} 
\caption{Comparison of DWMRPM and two deep-learning approaches on IMD
	dataset for a randomly coordinates situated
	at 25$^\circ$25'N, 75$^\circ$50'E. The results are for the months
	of June, July,
	August and September of 23 years (June 1996 to September 2018)
	(a) Prediction
	results of MLP, (b)
	Prediction results of one dimensional CNN and, (d) Prediction results of the
	proposed DWMRPM.}
\label{fig:ComparisonIMD}
\end{figure*}

To establish the significance of present work, we
compare the results of our model with the baseline
	approaches separately using the IMD gridded dataset and WRD station
	dataset.Table~\ref{tab:ComparisonWRD} and Table~\ref{tab:ComparisonIMD} show the comparison of the proposed DWMRPM with
	other approaches in the prediction of monsoon rainfall for the
	months of June, July, August and September on WRD and IMD datasets
	respectively. Overall accuracy of the model in the prediction of
	rainfall for the monsoon months is also given. Qualitative
	analysis for the comparison on different datasets is shown in
	Figure~\ref{fig:ComparisonWRD} and Figure~\ref{fig:ComparisonIMD}

\begin{table}
	\caption{Comparison of the proposed DWMRPM with one Dimensional
	Convolutional Neural Networks (1-DCNN) and Multilayer Perceptron
	(MLP) on WRD dataset using for each month of the monsoon season.}
	\label{tab:ComparisonWRD}
\begin{tabular}{lllllll}
	\hline
		\multirow{2}{*}{Month} & \multicolumn{2}{c}{MLP} &
		\multicolumn{2}{c}{1-DCNN} & \multicolumn{2}{c}{DWMRPM} \\
		\cline{2-7}
		& RMSE & MAE & RMSE & MAE & RMSE & MAE \\
		\hline
June & 7.0382  & 5.8610  & 7.7567  & 5.2118 &  6.550 & 4.550  \\
July & 12.3831 & 9.2600 & 14.2568 & 10.3249 &  11.0974 & 8.7081  \\
August & 15.4046 & 12.3992 & 15.4564 & 10.4677 &  13.7013 & 10.4781  \\
September & 8.1199 & 9.5221 & 7.9481 & 5.9679 & 6.5770 & 5.0796  \\	
Overall Accuracy & 10.2014 & 7.0106 & 11.8901 & 7.9931 & 9.9637 & 7.2052  \\
\hline

\end{tabular}
\end{table}

\begin{table}
	\caption{Comparison of the proposed DWMRPM with one Dimensional
	Convolutional Neural Networks (1-DCNN) and Multi-layer Perceptron
	(MLP) on IMD dataset using for each month of the monsoon season.}
	\label{tab:ComparisonIMD}
\begin{tabular}{lllllll}
\hline
		\multirow{2}{*}{Month} & \multicolumn{2}{c}{MLP} &
		\multicolumn{2}{c}{1-DCNN} & \multicolumn{2}{c}{DWMRPM} \\
		\cline{2-7}
		& RMSE & MAE & RMSE & MAE & RMSE & MAE \\
		\hline
June & 6.8156 & 5.7843  & 5.9660 & 5.0429 & 4.0239 & 3.1371  \\
July & 12.9685 & 9.9012 & 12.7962  & 8.5378 & 11.7024 & 7.8652 \\
August & 13.8754 & 12.8701 & 13.3874  & 12.6042 & 12.6878 & 12.1112  \\
September & 6.5700 & 5.8955 & 4.6474 & 4.52739 & 3.8953 & 4.0184 \\	
Overall Accuracy  & 11.5009 & 8.4529 & 9.5039 & 4.6780 & 9.0598 & 4.2830 \\
\hline

\end{tabular}
\end{table}

%

\section{Conclusion and Future Work}
	\label{sec:Conclusion}
This paper has presented a deep and wide neural network based model for the prediction of  Rajasthan Summer Monsoon Rainfall (RSMR). Rainfall data is collected from Water Resource Department, Rajasthan and gridded data of resolution 0.25 X 0.25 degrees from Indian Meteorological Department (IMD). This model has the added advantage of exploiting the  benefits from both the interpolated gridded data set and the unbiased single point station data set as well. Results obtained by DWRM are compared with baseline approaches like MLP and CNN. It is observed that for RSMR, the deep and wide model works better than other approaches.
In future we may apply similar technique for the prediction of summer monsoon rainfall in other states in India as well as abroad. We plan to add more number of rainfall indicators and explore the possibilities of improving the accuracy of the current method.
%
%
%
%
%

\section{Acknowledgments}

This work is in collaboration with Water Resources, Government of Rajasthan. We are thankful to Indian Meteorological Department (IMD) and Special Project Monitoring Unit, National Hydrology Project, Water Resources Rajasthan, Jaipur, India for providing us the Rainfall data for this study.

\section{Declaration}
	\begin{itemize}[]
		\item[] \textbf{Funding:} Not Applicable
		\item[] \textbf{Conflicts of interest/Competing
			interests:} The authors certify that they have NO affiliations with or involvement in any organization or entity with any financial interest or non-financial interest in the subject matter or materials discussed in this manuscript.
		\item[] \textbf{Availability of data and material:}
			Available on request.

	\end{itemize}

\bibliographystyle{spbasic}
\bibliography{refs}

\end{document}